\documentclass[runningheads]{llncs}
\usepackage{graphicx}
\usepackage{amssymb} %for \checkmark
\usepackage[colorlinks=true,linkcolor=blue,anchorcolor=blue,citecolor=blue,urlcolor=blue]{hyperref}

\begin{document}
\title{Multi-branch and Multi-scale Attention Learning for Fine-Grained Visual Categorization}
\titlerunning{Multi-branch and Multi-scale Attention Learning for FGVC}
\author{Fan Zhang\inst{}\orcidID{0000-0003-2504-4651} \and
Meng Li\inst{}\orcidID{0000-0002-1543-5029} \and
Guisheng Zhai\inst{}\orcidID{0000-0001-9901-010X}\and
Yizhao Liu\inst{}\orcidID{0000-0002-9549-7317}}
\authorrunning{F. Zhang et al.}
\institute{School of MEIE , China University of Mining and Technology (Beijing), Beijing, China \\
\email{sqt1800407119@student.cumtb.edu.cn}}
\maketitle              % typeset the header of the contribution
\begin{abstract}
ImageNet Large Scale Visual Recognition Challenge (ILSVRC) is one of the most authoritative academic competitions in the field of Computer Vision (CV) in recent years. But applying ILSVRC's annual champion directly to fine-grained visual categorization (FGVC) tasks does not achieve good performance. To FGVC tasks, the small inter-class variations and the large intra-class variations make it a challenging problem. Our attention object location module (AOLM) can predict the position of the object and attention part proposal module (APPM) can propose informative part regions without the need of bounding-box or part annotations. The obtained object images not only contain almost the entire structure of the object, but also contains more details, part images have many different scales and more fine-grained features, and the raw images contain the complete object. The three kinds of training images are supervised by our multi-branch network. Therefore, our multi-branch and multi-scale learning network(MMAL-Net) has good classification ability and robustness for images of different scales. Our approach can be trained end-to-end, while provides short inference time. Through the comprehensive experiments demonstrate that our approach can achieves state-of-the-art results on CUB-200-2011, FGVC-Aircraft and Stanford Cars datasets. Our code will be available at \url{https://github.com/ZF1044404254/MMAL-Net}
\end{abstract}
\section{Introduction}
\begin{figure}[h]
\includegraphics[width=\textwidth]{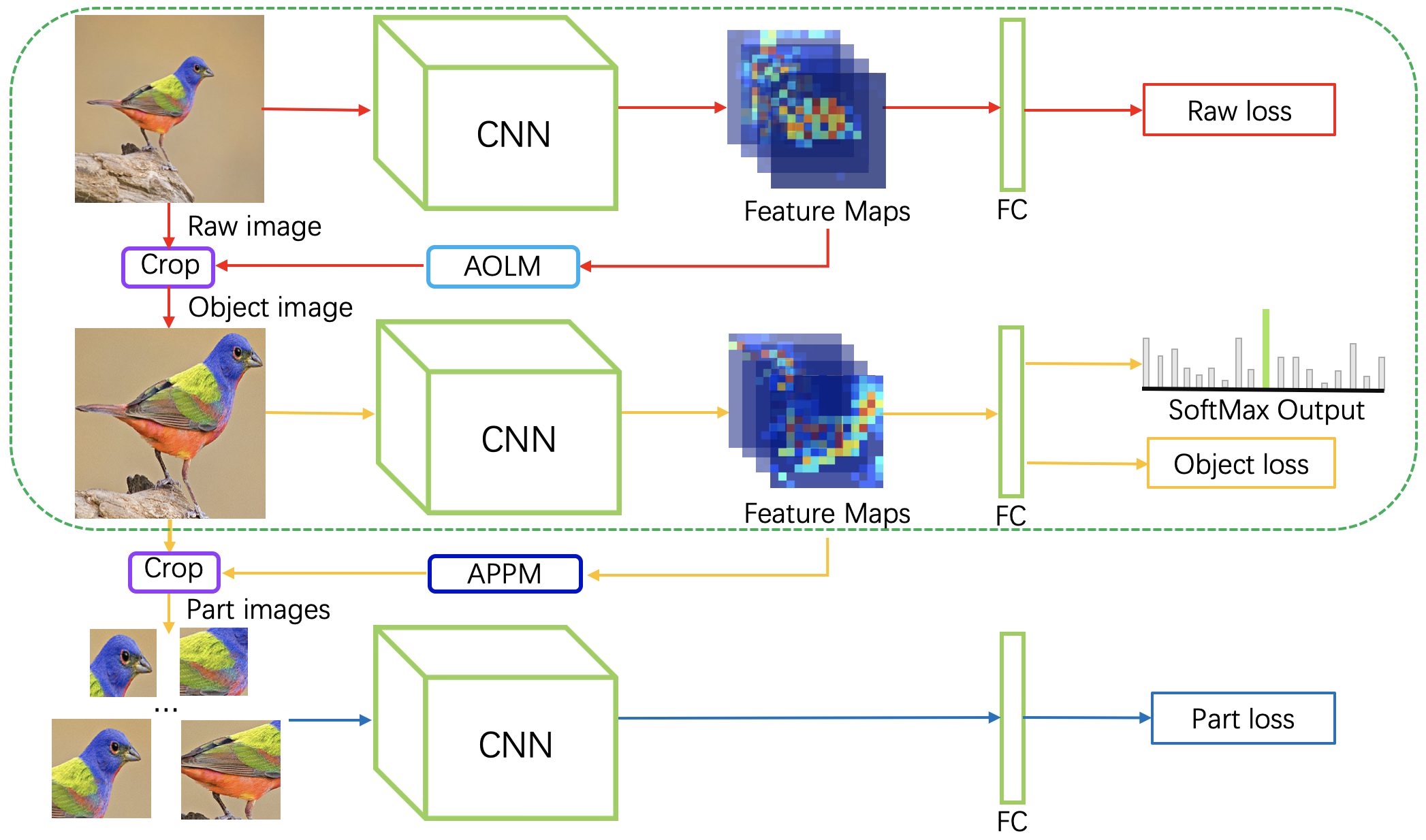}
\caption{The overview of our proposed MMAL-Net in the training phase, The red branch is raw branch, the orange branch is object branch, and the blue branch is part branch. In the dotted green box is the network structure for the test phase. The CNN (Convolutional Neural Networks) and FC (Fully Connection) layer of the same color represent parameter sharing. Our multi-branch all use cross entropy loss as the classification loss.} \label{fg1}
\end{figure}

% \subsection{A Subsection Sample}
How to tell a dog's breed? This is a frequently asked question because dogs have similar characteristics. The FGVC direction of CV research focuses on such issues, and it is also called sub-category recognition. In recent years, it is a very popular research topic in CV, pattern recognition and other fields. It's purpose is to make a more detailed sub-class division for coarse-grained large categories (e.g. classifying bird species \cite{WahCUB_200_2011}, aircraft models \cite{maji2013fine}, car models \cite{KrauseStarkDengFei-Fei_3DRR2013}, etc.).\\
\indent Many papers \cite{lin2015bilinear,zhang2016spda,wei2016mask,lam2017fine} have shown that  the key to fine-grained visual categorization tasks lies in developing effective methods to accurately identify informative regions in an image. They leverage the extra annotations of bounding box and part annotations to localize significant regions. However, obtaining such dense annotations of bounding box and part annotations is labor-intensive, which limits both scalability and practicality of real-world fine-grained applications. Other methods \cite{zhao2017diversified,zheng2017learning,yang2018learning,zheng2019looking} use an unsupervised learning scheme to localize informative regions. They eliminate the need for the expensive annotations, but how to focus on the right areas and use them is still worth investigating.\\
\indent An overview of our MMAL-Net is shown in Fig. \ref{fg1}. Our method has three branches in training phase,  whose raw branch mainly studies the overall characteristics of the object, and AOLM needs to obtain the object's bounding box information with the help of the feature maps of the raw image from this branch. As the input of object branch, the finer scale of object image is very helpful for classification, because it contains the structural features of the target as well as the fine-grained features. Then, APPM proposes several part regions with the most discrimency and less redundancy according to the feature maps of object image. The part branch sends the part image clipped from the object image to the network for training. It enables the network to learn fine-grained features of different parts in different scales. Unlike RA-CNN \cite{fu2017look}, the parameters of CNN and FC in our three branches are shared. Therefore, through the common learning process of the three branches, the trained model has a good classification ability for different scales and parts of object. In the testing phase, unlike RA-CNN \cite{fu2017look} and NTS-Net \cite{yang2018learning} need to calculate the feature vectors of the multiple part region images and then concat these vectors for classification. After our repeated experiments, the best classification performance is simply obtained by the result of object branch, so our approach can reduce some calculations and inference time while achieving high accuracy.\\
\indent Our main contributions can be summarized as follows: 
\begin{itemize}
\item[$\bullet$]Our multi-branch network can be trained end-to-end and learn object's discriminative regions for recognition effectively.
\item[$\bullet$]The AOLM does not increase the number of parameters, so we do not need to train a proposal network like RA-CNN \cite{fu2017look}. The accuracy of object localization is achieved by only using category labels.
\item[$\bullet$]We present an attention part proposal method(APPM) without the need of part annotations. It can select multiple ordered discriminative part images, so that the model can effectively learn different scales parts's fine-grained features.
\item[$\bullet$]State-of-the-art performances are reported on three standard benchmark datasets, where our method stable outperforms the state-of-the-art methods and baselines.
\end{itemize}
\section{Related works}
In the past few years, the accuracy of benchmark on open datasets has been improved based on deep learning and fine-grained classification methods. They can be classified as follows: 1) By end-to-end feature encoding; 2) By localization-classification subnetworks. 
\subsection{By End-to-End Feature Encoding}
This kind of method directly learns a more discriminative feature representation by developing powerful deep models for fine-grained recognition. The most representative method among them is Bilinear-CNN \cite{lin2015bilinear}, which represents an image as a pooled outer product of features derived from two bilinear models, and thus encodes higher order statistics of convolutional activations to enhance the mid-level learning capability. Thanks to its high model capacity, it achieves clear performance improvement on a wide range of visual tasks. However, the high dimensionality of bilinear features still limits its further generalization. In order to solve this problem, \cite{gao2016compact,cui2017kernel} try to aggregate low-dimensional embeddings by applying tensor sketching. They can reduce the dimensionality of bilinear features and achieve comparable or higher classification accuracy.
\subsection{By Localization-Classification Subnetworks}
This kind of method trains a localization subnetwork with supervised or weakly supervised to locates key part regions. Then the classification subnetwork uses the information of fine-grained regions captured by the localization subnetwork to further enhance its classification capability. Earlier works \cite{zhang2016spda,wei2016mask,lam2017fine} belong to full supervision method depend on more than image-level annotations to locate semantic key parts. \cite{lam2017fine} trained a region proposal network to generate proposals of informative image parts and concatenate multiple part-level features as a whole image representation toward final fine-grained recognition. However, maintaining such dense part annotations increases additional location labeling cost.
Therefore, \cite{zhao2017diversified,zheng2017learning,yang2018learning,zheng2019looking} take advantage of attention mechanisms to avoid this problem. There is no need of bounding box annotations and part annotations except image-level annotation.
\section{Method}
\subsection{Attention Object Location Module (AOLM)}
\begin{figure}
\includegraphics[width=\textwidth]{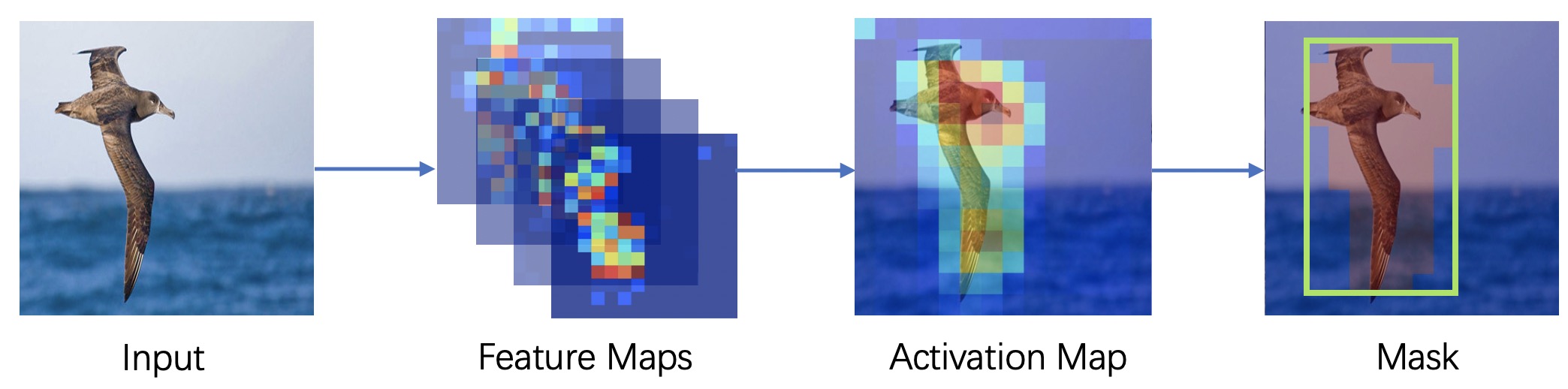}
\caption{The pipeline of the AOLM, we first get an activation map by aggregating the feature maps in the channel dimension, then obtain a bounding box according to activation map.} \label{fg2}
\end{figure}

This method was inspired by SCDA \cite{wei2017selective}. SCDA uses a pre-trained model to extract image features for fine-grained image retrieval tasks. We improve it's positioning performance as much as possible through some measures. At the first, we describe the process of generating object location coordinates by processing the CNNs feature map as the Fig. \ref{fg2} illustrated. 

We use  $ F \in R^{C \times H \times W} $ to represent feature maps with $ C $ channels and spatial size $ H \times W $ output from the last convolutional layer of an input image $ X $ and $ f_i $ is the i-th feature map of the corresponding channel. As shown in Equ \ref{eq1}, 
\begin{equation}
A =  \sum\limits_{i=0}^{C-1}f_i \label{eq1}
\end{equation}
activation map $ A $ can be obtained by aggregating the feature maps $ F $. We can visualizes where the deep neural networks focus on for recognition simply and locates the object regions accurately from A. As shown in Equ \ref{eq2}, $ \overline{a} $ is the mean value of $ A $.
\begin{equation}
\label{eq2}
\overline{a} =  \frac{\sum_{x=0}^{W-1}{\sum_{y=0}^{H-1}}A{(x, y)}}{H \times W}
\end{equation}
$ \overline{a} $ is used as the threshold to determine whether the element at that position in $ A $ is a part of object, and $ (x, y) $ is a particular position in a $ H \times W $ activation map. Then we initially obtained a coarse mask map $ \widetilde{M}_{conv\_5c} $ from the last convolutional layer  $ Conv\_5c $ of ResNet-50 \cite{he2016deep} according to Equ \ref{eq3}.
\begin{equation}
\label{eq3}
\widetilde{M}_{(x,y)}=\left\{
\begin{array}{ll}
1 & \mbox{            if } A_{(x,y)} > \overline{a} \\
0 & \mbox{            otherwise }
\end{array}
\right.
\end{equation}
On the basis of the experimental results, we find that the object is often in the largest connected component of $ \widetilde{M}_{conv\_5c} $, so the smallest bounding box containing the largest connected area is used as the result of our object location. Only using a pre-trained VGG16 \cite{simonyan2014very} in SCDA \cite{wei2017selective} achieved better position accuracy, but our pre-trained ResNet-50 does not reach a similar accuracy rate and dropped significantly. So we use the training set to train ResNet-50 for improving object location accuracy and experiments in subsection 4.5 verify the effectiveness of this approach. Then, inspired by \cite{wei2017selective} and \cite{long2015fully} the performance of their methods all benefit from the ensembel of Multiple layers. So we get the activation map of the output of $ Conv\_5b $ according to Equ \ref{eq1}, which is one block in front of $ Conv\_5c $. Then we can get $ \widetilde{M}_{conv\_5b} $ according to Equ \ref{eq3}, and finally we can get a more accurate mask M after taking the intersection of $ \widetilde{M}_{conv\_5c} $ and $ \widetilde{M}_{conv\_5b} $. As shown in Equ \ref{eq4}.
\begin{equation}
\label{eq4}
M =  \widetilde{M}_{conv\_5c} \cap \widetilde{M}_{conv\_5b}
\end{equation}
Subsequent experimental results prove the effectiveness of these approaches to improve object location accuracy. This improved weakly supervised object location method can achieve higher localization accuracy than ACOL \cite{zhang2018adversarial}, ADL \cite{choe2019attention} and SCDA \cite{wei2017selective}, without adding trainable parameters.
\subsection{Attention Part Proposal Module(APPM)}
\begin{figure}
\includegraphics[width=\textwidth]{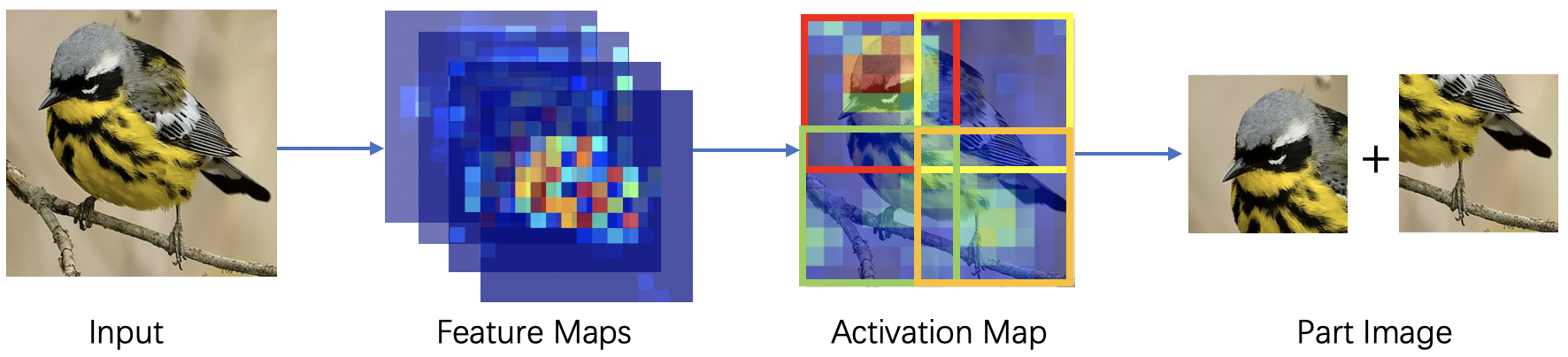}
\caption{The simple pipeline of the APPM. We use red, orange, yellow, green colors to indicate the order of windows' $ \overline{a}_w $.} \label{fg3}
\end{figure}

Although AOLM can achieve higher localization accuracy, but there are some positioning results are part of the object. We improve the robustness of the model to this situation through APPM and part branch. It will be demonstrated in the next section. By observing the activation map $ A $, we find that the area with high activation value of the activation map are often the area where the key part are located, such as the head area in the example. Using the idea of sliding window in object detection to find the windows with information as part images. Moreover, we implemented the traditional sliding window approach with full convolutional network to reduce the amount of calculation, just like Overfeat \cite{sermanet2013overfeat} gets the feature map of different windows from the feature map output from the previous branch. Then we aggregate each window's activation map $ A_w $ in the channel dimension and get its activation mean value $ \overline{a}_w $ according to Equ \ref{eq5}, 
\begin{equation}
\label{eq5}
\overline{a}_w =  \frac{\sum_{x=0}^{W_w-1}{\sum_{y=0}^{H_w-1}}A_w{(x, y)}}{H_w \times W_w}
\end{equation}
$ H_w $ , $ W_w $ are the height and width of a window's feature map. We sort by the $ \overline{a}_w $ value of all windows. The larger the $ \overline{a}_w $ is, the larger the informativeness of this part region is, as shown in Fig. \ref {fg3}. However, we cannot directly select the first few windows, because they are often adjacent to the largest $ \overline{a}_w $ windows and contain approximate the same part, but we hope to select as many different parts as possible. In order to reduce region redundancy, we adopt non-maximum suppression (NMS) to select a fixed number of windows as part images with different scales. By visualizing the output of this module in Fig. \ref {fg4}, it can be seen that this method proposed some ordered, different importance degree part regions.
\subsection{Architecture of MMAL-Net}
In order to make the model fully and effectively learn the images obtained through AOLM and APPM. During the training phase, we construct a three branches network structure consisting of raw branch, object branch, and part branch, as shown in fig. \ref {fg1}. The three branches share a CNN for feature extraction and a FC layer for classification. Our three branches all use cross entropy loss as the classification loss. As shown in Equ \ref{eq6}, \ref{eq7}, and \ref{eq8}, respectively.
\begin{equation}
\label{eq6}
L_{raw} = -\log (P_r(c))
\end{equation}
\begin{equation}
\label{eq7}
L_{object} = -\log (P_o(c))
\end{equation}
\begin{equation}
\label{eq8}
L_{parts} = -\sum\limits_{n=0}^{N-1}\log (P_{p_(n)}(c))
\end{equation}
Where $ c $ is the ground truth label of the input image, $ P_r $, $ P_o $ are the category probabilities of the last softmax layer output of the raw branch and object branch, respectively , $ P_{p_(n)} $ is the output of the softmax layer of the part branch corresponding to the $n$th part image, $ N $ is the number of part images. The total loss is defined as:
\begin{equation}
\label{eq9}
L_{total} = L_{raw}+ L_{object}+ L_{parts}
\end{equation}
The total loss is the sum of the losses of the three branches, which work together to optimize the performance of the model during backpropagation. It enables the final convergent model to make classification predictions based on the overall structural characteristics of the object or the fine-grained characteristics of a part. The model has good object scale adaptability, which improves the robustness in the case of inaccurate AOLM localization. During the testing phase, we removed the part branch so as to reduce a large amount of calculations, so our method will not take too long to predict in practical applications. Due to our reasonable and efficient framework. MMAL-Net can achieves the state-of-the-art performance so far.
\section{Experiments}
\subsection{Datasets}
These three datasets are widely used as benchmarks for fine-grained classification(shown in Table~\ref {tb1}). In our experiments, we only use the image classification labels provided by these datasets. 
\begin{table}
\centering
\caption{Introduction of the three datasets used in this paper.}\label{tb1}
\begin{tabular}{|c|c|c|c|c|}
\hline
Datasets & Object & Class & Train & Test\\
\hline
CUB-200-2011(CUB) \cite{WahCUB_200_2011} & Bird & 200 & 5994 &  5794\\
FGVC-Aircraft(AIR) \cite{maji2013fine} & Aircraft & 100  & 6667 &  3333\\
Stanford Cars(CAR) \cite{KrauseStarkDengFei-Fei_3DRR2013} & Car &  196  & 8144 &  8041\\
\hline
\end{tabular}
\end{table}
\subsection{Implementation Details}
In all our experiments, we first preprocess images to size $448 \times 448$ to get input image for raw branch and object branch. The object image is also scaled into $448 \times 448$ , but all part images are resized to $224 \times 224$ for part branch. We construct windows with three broad categories of scales: \{[$ 4 \times 4 $, $ 3 \times 5 $], [$ 6 \times 6 $, $ 5 \times 7 $], [$ 8 \times 8 $, $ 6 \times 10 $, $ 7 \times 9 $, $ 7 \times 10 $]\} for activation map of $14 \times 14$ size, and the number of a raw image's part images is $ N = 7 $ , among them $ N_1 = 2, N_2 = 3, N_3 = 2 $. $ N_1$, $ N_2 $ and $N_3 $ are the number of three broad categories of scales windows mentioned above. ResNet-50 \cite{he2016deep} pre-trained on ImageNet is used as the backbone of our network structure. During training and testing, we do not use any other annotations other than image-level labels. Our optimizer is SGD with the momentum of 0.9 and the weight decay of 0.0001, and a mini-batch size of 6 on a Tesla P100 GPU. The initial learning rate is 0.001 and multiplied by 0.1 after 60 epoch. We use Pytorch as our code-base.
\subsection{Performance Comparison}
 We compared the baseline methods mentioned above on three commonly used fine-grained classification datasets. The experimental results are shown in the Table~\ref {tb2}. By comparison, we can see that our method achieves the best accuracy currently available on these three datasets.
\begin{table}
\centering
\caption{Comparison results on three common datasets. Train Anno. represents using bounding box or part annotations in training.}\label{tb2}
\begin{tabular}{|c|c|c|c|c|}
	\hline
	Methods & Train Anno. & CUB & AIR & CAR \\
	\hline
	ResNet-50 & & & & \\
	\hline
	Bilinear-CNN \cite{lin2015bilinear} &  & 84.1 & 84.1 & 91.3 \\
	SPDA-CNN \cite{zhang2016spda} & \checkmark & 85.1 & - & - \\
	KP \cite{cui2017kernel} & & 86.2 & 86.9 & 92.4 \\
	RA-CNN \cite{fu2017look} & & 85.3 & - & 92.5 \\
	MA-CNN \cite{zheng2017learning} & & 86.5 & 89.9 & 92.8 \\
	OSME+MAMC \cite{sun2018multi} & & 86.5 & - & 93.0 \\
	PC \cite{dubey2018pairwise} & & 86.9 & 89.2 & 92.9 \\
	HBP \cite{yu2018hierarchical} & & 87.1 & 90.3 & 93.7 \\
	Mask-CNN \cite{wei2016mask} & \checkmark & 87.3 & - & - \\
	DFL-CNN \cite{wang2018learning} & & 87.4 & 92.0 & 93.8 \\
	HSnet \cite{lam2017fine} & \checkmark & 87.5 & - & - \\
	NTS-Net \cite{yang2018learning} & & 87.5 & 91.4 & 93.9 \\
	MetaFGNet \cite{zhang2018fine} & & 87.6 & - & - \\
	DCL \cite{chen2019destruction} & & 87.8 & \textbf{92.2} & \textbf{94.5} \\
	TASN \cite{zheng2019looking} & & \textbf{87.9} & - & 93.8 \\
	\hline
	Ours & & \textbf{89.6} & \textbf{94.7} & \textbf{95.0} \\
	\hline
\end{tabular}
\end{table}
\subsection{Ablation Studies}
The ablation study is performed on the CUB dataset. Without adding any of our proposed methods, the ResNet-50 \cite{he2016deep} obtained an accuracy of 84.5\% under the condition that the input image resolution is $ 448 \times 448 $.  In order to verify the rationality of the training structure of our three branches, we remove the object branch and part branch respectively. After removing the object branch, the best accuracy is 85.0\% from the raw branch, a drop of 4.6\%. This proves the great contribution of AOLM and object branch to improve the classification accuracy. After removing the part branch, the best accuracy is 87.3\% from the object branch, down 2.3\% . The results of the experiment show that part branch and APPM can improve the robustness of the model when facing AOLM unstable positioning results. The above experiment has shown that the three branches of our method all have a significant contribution to the final accuracy. 
\subsection{Object Localization Performance}
Percentage of Correctly Localized Parts (PCP) metric is the percentage of predicted boxes that are correctly localized with more than 50\% IOU with the ground-truth bounding boxes. On the CUB dataset, the best result of AOLM in terms of the PCP metric for object localization is 85.1\%. AOLM clearly exceeds SCDA's \cite{wei2017selective} 76.8\% and the recent weakly supervised object location methods ACOL’s \cite{zhang2018adversarial} 46.0\% and  ADL’s \cite{choe2019attention} 62.3\%. As shown in Table~\ref {tb3}, the ensemble of multiple layers significantly improves object location accuracy. Through the experiment, we find that the object location accuracy using the pre-trained model directly is 65.0\%. However, it rise to 85.1\% after one iteration of training. And as the training progressed, CNN paid more and more attention to the most discerning region to improve classification accuracy which leads to localization accuracy drops to 71.1\%. Even so, due to part branch and APPM make model has good adaptability to object's scale, it still achieved excellent classification performance.
\begin{table}
\centering
\caption{Object localization accuracy on CUB-200-2011.}\label{tb3}
\begin{tabular}{|c|c|}
	\hline
	Methods & localization Accuracy \\ 
	\hline
	ACOL \cite{zhang2018adversarial} & 46.0\% \\ 
	ADL \cite{choe2019attention} & 62.3\% \\ 
	SCDA \cite{wei2017selective} &  \textbf{76.8\%} \\ 
	\hline
	AOLM(conv\_5c) & 82.2\% \\ 
	AOLM(conv\_5b \& conv\_5c) & \textbf{85.1\%} \\
	\hline
\end{tabular}
\end{table}
\subsection{Model and Inference Time Complexity}
Unlike RA-CNN \cite{fu2017look} and NTS-Net \cite{yang2018learning}, the former has three branches with independent parameters and needs to train a subnetwork to propose finer scale images and the latter needs to train a navigator network to propose regions with large amount of information (such as different body parts of birds). Our MMAL-Net has some advantages in terms of parameter volume over them. First, the three branches parameters are shared, and secondly the AOLM and APPM modules do not require training, do not add additional parameter amounts. Thirdly their calculation amount is relatively smaller. Finally, better classification performance is achieved by MMAL-Net. Compared with the ResNet-50 baseline, our method yields a significantly better result (+4.1\%) with the same parameter volume. As for inference time, RA-CNN needs to calculate the output of three branches and fuse them; NTS-Net needs to extract and fuse the 4 proposal local image features of the input image. Above reasons make their inference time relatively longer and our method has a shorter inference time. It only needs to calculate the output of the first two branches and does not need to fuse them, because the classification results are based on the output of the second branch (object branch). For more accurate comparison, we conducted an inference time test on Tesla P100 and the input image size is  $448 \times 448$. The inference time of MMAL-Net and NTS-Net  are summarized as follows: the running time on an image of NTS-Net is about 5.61 ms, and ours method’s running time on an image is about 1.80 ms. Based on the above analysis, lower model and inference time complexity both add extra practical value to our proposed method.
\subsection{Visualization of Object and Part Regions}
\begin{figure}
\centering
\includegraphics[width=10cm]{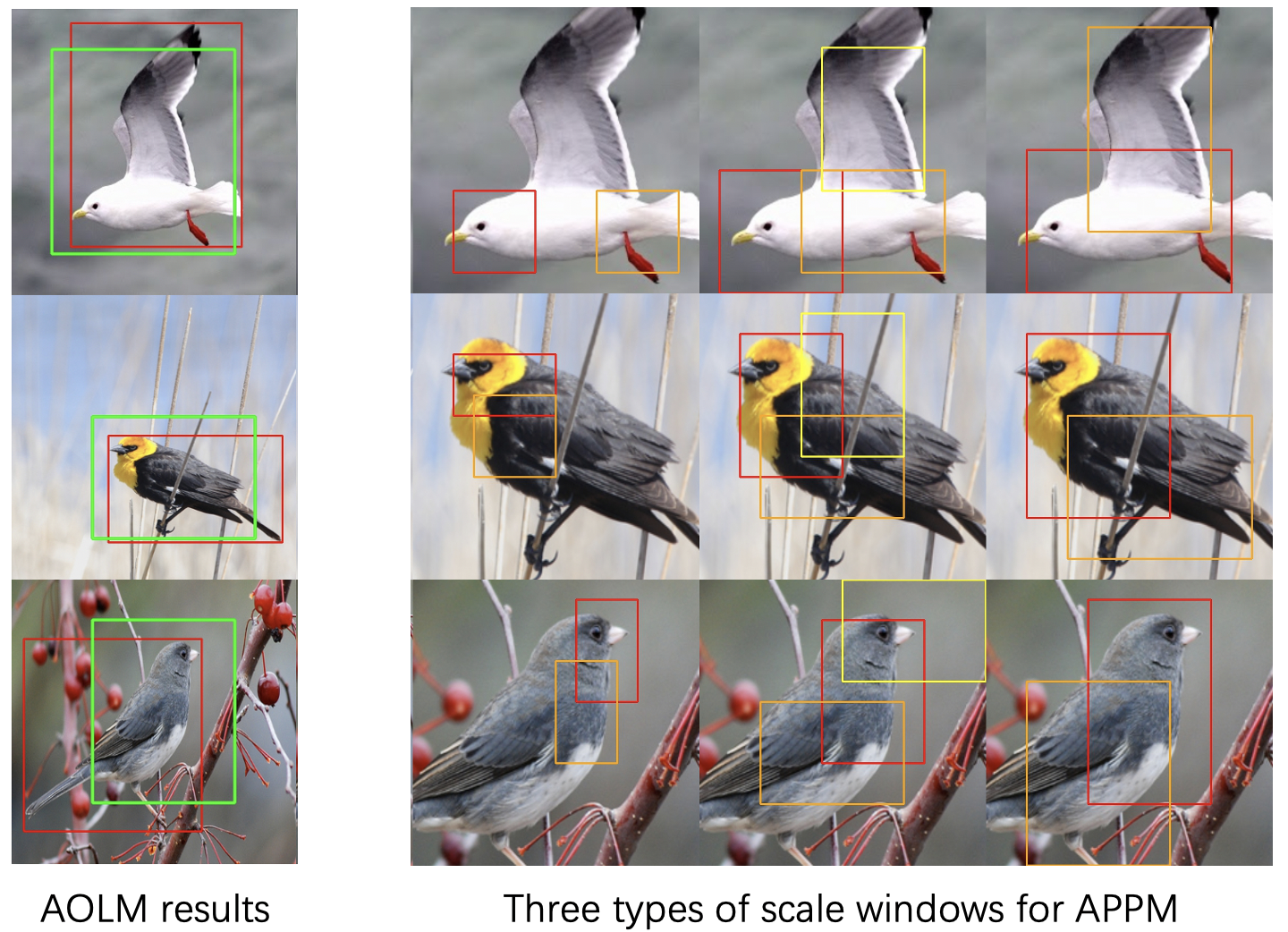}
\caption{Visualization of object and part regions.} \label{fg4}
\end{figure}

In order to visually analyze the areas of concern for our AOLM and APPM, we draw the object's bounding boxes and part regions proposed by AOLM and APPM in Fig. \ref {fg4}. In the first column, we use red and green rectangles to denote the ground truth and predicted bounding box in raw image. It is very helpful for classification that the positioned areas of objects often cover an almost complete object. In columns two through four, we use red, orange, yellow, and green rectangles to represent the regions with the highest average activation values in different scales proposed by APPM, with red rectangle denoting the highest one.  Fig. \ref {fg4} conveys that the proposed area does contain more fine-grained information and the order is more reasonable on the same scale, which are very helpful for model's robustness to scale. We can find that the most discriminative regions of birds are head firstly , then is body, which is similar to human cognition.
\section{Conclusion}
In this paper, we propose an effective method for fine-grained classification without bounding box or part annotations. The multi-branch structure can make full use of the images obtained by AOLM and APPM to achieve excellent performance. Our algorithm is end-to-end trainable and achieves state-of-the-art results in CUB-200-2001 \cite{WahCUB_200_2011}, FGVC Aircraft \cite{maji2013fine} and Stanford Cars \cite{KrauseStarkDengFei-Fei_3DRR2013}  datasets.The future work is how to set the number and size of windows adaptively to further improve the classification accuracy. \\

%For citations of references, we prefer the use of square brackets
%and consecutive numbers. Citations using labels or the author/year
%convention are also acceptable. The following bibliography provides
%a sample reference list with entries for journal
%articles~\cite{ref_article1}, an LNCS chapter~\cite{ref_lncs1}, a
%book~\cite{ref_book1}, proceedings without editors~\cite{ref_proc1},
%and a homepage~\cite{ref_url1}. Multiple citations are grouped
%\cite{ref_article1,ref_lncs1,ref_book1},
%\cite{ref_article1,ref_book1,ref_proc1,ref_url1}.

\bibliographystyle{splncs04unsort}
\bibliography{refs}

\begin{thebibliography}{10}
\providecommand{\url}[1]{\texttt{#1}}
\providecommand{\urlprefix}{URL }
\providecommand{\doi}[1]{https://doi.org/#1}

\bibitem{WahCUB_200_2011}
Wah, C., Branson, S., Welinder, P., Perona, P., Belongie, S.: {The Caltech-UCSD
  Birds-200-2011 Dataset}. Tech. Rep. CNS-TR-2011-001, California Institute of
  Technology (2011)

\bibitem{maji2013fine}
Maji, S., Rahtu, E., Kannala, J., Blaschko, M., Vedaldi, A.: Fine-grained
  visual classification of aircraft. arXiv preprint arXiv:1306.5151  (2013)

\bibitem{KrauseStarkDengFei-Fei_3DRR2013}
Krause, J., Stark, M., Deng, J., Fei-Fei, L.: 3d object representations for
  fine-grained categorization. In: 4th International IEEE Workshop on 3D
  Representation and Recognition (3dRR-13). Sydney, Australia (2013)

\bibitem{lin2015bilinear}
Lin, T.Y., RoyChowdhury, A., Maji, S.: Bilinear cnn models for fine-grained
  visual recognition. In: CVPR. pp. 1449--1457 (2015)

\bibitem{zhang2016spda}
Zhang, H., Xu, T., Elhoseiny, M., Huang, X., Zhang, S., Elgammal, A., Metaxas,
  D.: Spda-cnn: Unifying semantic part detection and abstraction for
  fine-grained recognition. In: Proceedings of the IEEE Conference on Computer
  Vision and Pattern Recognition. pp. 1143--1152 (2016)

\bibitem{wei2016mask}
Wei, X.S., Xie, C.W., Wu, J.: Mask-cnn: Localizing parts and selecting
  descriptors for fine-grained image recognition. arXiv preprint
  arXiv:1605.06878  (2016)

\bibitem{lam2017fine}
Lam, M., Mahasseni, B., Todorovic, S.: Fine-grained recognition as hsnet search
  for informative image parts. In: CVPR. pp. 2520--2529 (2017)

\bibitem{zhao2017diversified}
Zhao, B., Wu, X., Feng, J., Peng, Q., Yan, S.: Diversified visual attention
  networks for fine-grained object classification. IEEE Transactions on
  Multimedia  \textbf{19}(6),  1245--1256 (2017)

\bibitem{zheng2017learning}
Zheng, H., Fu, J., Mei, T., Luo, J.: Learning multi-attention convolutional
  neural network for fine-grained image recognition. In: ICCV. pp. 5209--5217
  (2017)

\bibitem{yang2018learning}
Yang, Z., Luo, T., Wang, D., Hu, Z., Gao, J., Wang, L.: Learning to navigate
  for fine-grained classification. In: ECCV. pp. 420--435 (2018)

\bibitem{zheng2019looking}
Zheng, H., Fu, J., Zha, Z.J., Luo, J.: Looking for the devil in the details:
  Learning trilinear attention sampling network for fine-grained image
  recognition. In: CVPR. pp. 5012--5021 (2019)

\bibitem{fu2017look}
Fu, J., Zheng, H., Mei, T.: Look closer to see better: Recurrent attention
  convolutional neural network for fine-grained image recognition. In: CVPR.
  pp. 4438--4446 (2017)

\bibitem{gao2016compact}
Gao, Y., Beijbom, O., Zhang, N., Darrell, T.: Compact bilinear pooling. In:
  CVPR. pp. 317--326 (2016)

\bibitem{cui2017kernel}
Cui, Y., Zhou, F., Wang, J., Liu, X., Lin, Y., Belongie, S.: Kernel pooling for
  convolutional neural networks. In: CVPR. pp. 2921--2930 (2017)

\bibitem{wei2017selective}
Wei, X.S., Luo, J.H., Wu, J., Zhou, Z.H.: Selective convolutional descriptor
  aggregation for fine-grained image retrieval. IEEE T IMAGE PROCESS
  \textbf{26}(6),  2868--2881 (2017)

\bibitem{he2016deep}
He, K., Zhang, X., Ren, S., Sun, J.: Deep residual learning for image
  recognition. In: CVPR. pp. 770--778 (2016)

\bibitem{simonyan2014very}
Simonyan, K., Zisserman, A.: Very deep convolutional networks for large-scale
  image recognition. arXiv preprint arXiv:1409.1556  (2014)

\bibitem{long2015fully}
Long, J., Shelhamer, E., Darrell, T.: Fully convolutional networks for semantic
  segmentation. In: CVPR. pp. 3431--3440 (2015)

\bibitem{zhang2018adversarial}
Zhang, X., Wei, Y., Feng, J., Yang, Y., Huang, T.S.: Adversarial complementary
  learning for weakly supervised object localization. In: CVPR. pp. 1325--1334
  (2018)

\bibitem{choe2019attention}
Choe, J., Shim, H.: Attention-based dropout layer for weakly supervised object
  localization. In: CVPR. pp. 2219--2228 (2019)

\bibitem{sermanet2013overfeat}
Sermanet, P., Eigen, D., Zhang, X., Mathieu, M., Fergus, R., LeCun, Y.:
  Overfeat: Integrated recognition, localization and detection using
  convolutional networks. arXiv preprint arXiv:1312.6229  (2013)

\bibitem{sun2018multi}
Sun, M., Yuan, Y., Zhou, F., Ding, E.: Multi-attention multi-class constraint
  for fine-grained image recognition. In: ECCV. pp. 805--821 (2018)

\bibitem{dubey2018pairwise}
Dubey, A., Gupta, O., Guo, P., Raskar, R., Farrell, R., Naik, N.: Pairwise
  confusion for fine-grained visual classification. In: ECCV. pp. 70--86 (2018)

\bibitem{yu2018hierarchical}
Yu, C., Zhao, X., Zheng, Q., Zhang, P., You, X.: Hierarchical bilinear pooling
  for fine-grained visual recognition. In: Proceedings of the European
  conference on computer vision (ECCV). pp. 574--589 (2018)

\bibitem{wang2018learning}
Wang, Y., Morariu, V.I., Davis, L.S.: Learning a discriminative filter bank
  within a cnn for fine-grained recognition. In: CVPR. pp. 4148--4157 (2018)

\bibitem{zhang2018fine}
Zhang, Y., Tang, H., Jia, K.: Fine-grained visual categorization using
  meta-learning optimization with sample selection of auxiliary data. In:
  Proceedings of the european conference on computer vision (ECCV). pp.
  233--248 (2018)

\bibitem{chen2019destruction}
Chen, S., Bai, Y., Zhang, W., Mei, T.: Destruction and construction learning
  for fine-grained image recognition. In: CVPR. pp. 5157--5166 (2019)

\end{thebibliography}
\end{document}